\documentclass[pdflatex,sn-nature]{sn-jnl}% Style for submissions to Nature Portfolio journals

\usepackage{graphicx}%
\usepackage{multirow}%
\usepackage{amsmath,amssymb,amsfonts}%
\usepackage{amsthm}%
\usepackage{mathrsfs}%
\usepackage[title]{appendix}%
\usepackage{xcolor}%
\usepackage{textcomp}% 
\usepackage{manyfoot}%
\usepackage{booktabs}%
\usepackage{algorithm}%
\usepackage{algorithmicx}%
\usepackage{algpseudocode}%
\usepackage{listings}%
\usepackage[english]{babel}
\usepackage{tabularx}
\usepackage{caption}
\usepackage{bm}
\usepackage{quantikz}
\graphicspath{{./Figures/}{./}}
\usepackage{lineno}
\begin{document}
%\linenumbers
\title[Quantum Generative Diffusion Model for Real-World Time Series]{Quantum Generative Diffusion Model for Real-World Time Series}

\author[1]{\fnm{Jack} \sur{Waller}}\email{k2407357@kingston.ac.uk}

\author[2]{\fnm{Filippo} \sur{Caruso}}\email{filippo.caruso@unifi.it}

\author[1]{\fnm{Dimitrios} \sur{Makris}}\email{D.Makris@kingston.ac.uk}

\author[3]{\fnm{Rajagopal} \sur{Nilavalan}}\email{Nila.Nilavalan@brunel.ac.uk}

\author*[1]{\fnm{Xing} \sur{Liang}}\email{X.Liang@kingston.ac.uk}

\affil*[1]{\orgdiv{School of Computer Science and Mathematics}, \orgname{Kingston University London}, \orgaddress{\city{London}, \postcode{KT1 2EE}, \country{U.K.}}}

\affil[2]{\orgdiv{Department of Physics and Astronomy}, \orgname{University of Florence}, \orgaddress{\city{Florence}, \postcode{Sesto Fiorentino 50019}, \country{Italy}}}

\affil[3]{\orgdiv{Department of Electronics and Electrical Engineering}, \orgname{Brunel University of London}, \orgaddress{\city{London}, \postcode{UB8 3PH}, \country{U.K.}}}

\abstract{Generative models have achieved remarkable success in data synthesis, though recent advances driven by increasing model scale have introduced challenges in computational cost and efficiency. Quantum machine learning offers a promising alternative, representing complex data distributions using compact, highly expressive models. Here, we propose QDiffusion-TS, the first quantum generative diffusion model for time series synthesis, and validate it on the IQM quantum processor. The framework extends a classical diffusion architecture by replacing feed-forward components within the denoising transformer with quantum neural networks, yielding a hybrid quantum transformer that reduces the number of trainable parameters in each replaced component by nearly three orders of magnitude. Evaluated on financial time series from Apple and Amazon, the model generates synthetic data that more accurately reproduces the real distributions, reducing Wasserstein distance by approximately 44\% relative to its classical counterpart across both datasets. In a downstream forecasting task, augmentation with the generated data improves predictive performance by up to 71\% in RMSE over a baseline trained solely on real data. These results show that quantum enhanced architectures can consistently match and frequently surpass classical performance with substantially fewer parameters, establishing a practical framework towards more efficient and scalable data-driven generative modelling.}


\maketitle

\section{Introduction}\label{sec1}
Generative machine learning models have experienced rapid advancement in recent years, achieving remarkable empirical performance across a wide range of data synthesis tasks \cite{31Sengar,32Ooi,33Fahrner}. Among these, image generation has progressed especially quickly, with variational autoencoders (VAEs) \cite{34Huang,35Daniel}, generative adversarial networks (GANs) \cite{63Goodfellow,36Gulrajani,37Karras}, and diffusion models \cite{61Ho,62Dickstein,41Rombach,38Dhariwal,39Kingma, 80Lu} now capable of producing samples that are highly realistic. Whilst progress has been considerable, these gains have been driven in large part by the increasing scale of model capacity and training data \cite{42Kaplan}. State-of-the-art models, such as DALL·E 3 \cite{66OpenAI} and Stable Diffusion \cite{41Rombach}, comprise billions of parameters \cite{65Ramesh,64Esser,43Ramesh,44Saharia}, raising concerns regarding computational cost, energy consumption, and further societal implications \cite{45Wynsberghe,46Crawford}.

Beyond synthesising images, generative models have also been applied to time series data \cite{48Brophy,49Li,50Yoon}, adapting similar architectures to capture temporal dependencies and reproduce the statistical properties of the underlying data. While sampling from simple distributions is relatively straightforward, accurately modelling complex, high dimensional distributions remains challenging. Financial time series provide a particularly demanding setting \cite{16Potluru}, requiring generative models to produce synthetic sequences that replicate the distribution of log-returns, preserve stylised facts, and maintain the broader statistical structure of real market data. A range of generative approaches have been previously explored in this context. Denoising autoencoders have been used to transfer distributional characteristics onto generated data, producing financial time series that differ sufficiently from purely stochastic simulations \cite{14Silva}. GAN-based methods have also been applied to generate realistic financial sequences, including frameworks designed to capture volatility structure \cite{23Wiese}, and cross-asset correlations \cite{17Masi}. Comparative studies across GANs, VAEs, and generative moment matching networks (GMMNs) suggest that although competitive performance can be achieved across architectures, VAE-based approaches tend to provide more stable and reliable results \cite{5Dogariu}. More recently, diffusion-based approaches have been introduced, including methods that operate on wavelet transformed representations of financial data for NASDAQ \cite{4Takahashi} and Japanese market datasets \cite{13Tanaka}. Score-based generative frameworks have further extended diffusion models by learning the data score function, enabling generation of data across varying temporal resolutions \cite{11Wang}.

Quantum machine learning (QML) has emerged as an extension of classical machine learning,
seeking to exploit the computational properties of quantum hardware through quantum algorithms to enhance model capabilities. Quantum neural networks (QNNs) employ parameterised quantum circuits (PQCs) as a learning framework \cite{52Benedetti,54Mitarai,53Schuld}, leveraging the exponential space state of quantum systems and entanglement between qubits to better represent complex data distributions. This enables expressive representations that can be achieved with significantly fewer trainable parameters than classical counterparts \cite{55Abbas}, as demonstrated across tasks such as graph classification \cite{87Qiao}. 

QNNs have been applied to generative learning in both fully quantum models, and hybrid settings where quantum components are embedded within otherwise classical models. A range of quantum generative architectures have been proposed, including Born machines that exploit shallow PQCs for generative tasks on near-term hardware \cite{88Benedetti}, and quantum generative adversarial networks (QGANs) \cite{82Lloyd}. Implementations of QGANs include models with both quantum generators and discriminators trained via quantum state fidelity objectives \cite{83Braccia, 84Braccia}, as well as entangling QGAN frameworks demonstrated on real quantum hardware \cite{57Niu}. Subsequent works have explored variations in circuit design, including compact quantum generators designed to reduce circuit depth \cite{58Chu}. Hybrid QGAN frameworks have also been applied to learning and loading probability distributions, including applications in financial derivative pricing \cite{86Zoufal}. More recently, quantum Wasserstein GANs have been applied to financial time series generation, utilising QNN-based generators alongside classical or quantum discriminators \cite{15Orlandi,22Dechant}.

This paper focuses on quantum diffusion models (QDMs), motivated by the success of classical diffusion architectures, which achieve state-of-the-art generative performance and have been shown to surpass GAN-based approaches in tasks such as image synthesis \cite{38Dhariwal}. QDMs were first proposed in \cite{1Parigi}, where quantum noise was leveraged to generate probability distributions that are intractable for classical processors, with the approach demonstrated on synthetic proof-of-concept distributions. In this setting, quantum features can be introduced into either the forward process or into the reverse denoising process through the integration of QNNs. Following the forward quantum noising direction, subsequent physics-inspired QDMs have harnessed the intrinsic noise of real quantum hardware as part of the diffusion process and applied it to image generation \cite{85Parigi}. Building on the original framework, the quantum denoising paradigm has been extended in a parallel line of work across a range of generative settings, including image synthesis on multiple datasets \cite{6Kolle} and architectures employing repeated circuit layers to enhance expressivity \cite{29Zhang}. Quantum latent diffusion models, in which classical autoencoders compress the data and QNNs operate within the reduced feature space, have also been explored \cite{30Falco}, including conditional variants for image generation tasks \cite{8Cacioppo},  and approaches combining vector quantisation with quantum autoencoders for quantum state generation \cite{81Li}. In addition, optimised quantum implicit neural networks (OQINN) have been proposed as alternative denoising models for image synthesis \cite{60Zhang}.

Despite rapid progress in both quantum machine learning and generative models, to our knowledge no study has applied a QDM to the synthesis of time series data. Here we introduce QDiffusion-TS, that is a QDM for time series generation that extends a classical diffusion architecture designed for temporal data by incorporating QNNs into the denoising process. The model is applied to real financial time series, which exhibit non-stationarity, heavy-tailed distributions, and complex temporal dependencies that are broadly representative of the challenges encountered across time series domains. In this formulation, QNNs (Fig. \ref{fig:qdiff}b) fully replace the feed-forward neural network (FFNN) components within the classical transformer, demonstrating how quantum circuits can be integrated into modern hybrid deep learning architectures within the constraints of near-term quantum hardware. The proposed approach is evaluated on financial time series data from Apple and Amazon, generating synthetic data that accurately reproduce the statistical distributions and stylised facts of the real data. QDiffusion-TS achieves consistently lower statistical distances across both datasets, reducing Wasserstein distance by approximately 44\% on average across features relative to its classical counterpart. To assess practical utility, the generated data is used to augment training datasets in a downstream forecasting task. This augmentation leads to substantial improvements, up to approximately 71\% in RMSE in predictive performance relative to a baseline trained solely on real data, with quantum and classical generated data yielding comparable peak forecasting accuracy. These findings are further supported by validation on quantum hardware, which not only outperforms classical baselines but also marginally exceeds the results obtained in simulation, indicating the feasibility of the proposed approach with current devices. A further key contribution of this work is the demonstration that integrating QNNs into existing large scale architectures can significantly reduce the number of trainable parameters, with each replaced component containing nearly three orders of magnitude fewer parameters. Despite this reduction in model complexity, the quantum model not only maintains the performance relative to the classical model, but in many cases, particularly in the reproduction of statistical distributions, surpasses it. These results highlight the potential of quantum enhanced architectures as a parameter efficient approach to generative modelling. As quantum hardware continues to scale, the significant parameter reductions demonstrated suggest that QDiffusion-TS provides a viable route towards hybrid models in which quantum components play progressively larger roles in generative architectures.

\section{Results}\label{sec2}
\subsection{QNN Preliminaries}
This section will give a brief introduction to QNNs, models that leverage quantum computing principles to augment classical neural networks. The structure of a QNN broadly resembles that of a classical neural network, with the key distinction that classical neurons are replaced with PQCs \cite{52Benedetti,54Mitarai,53Schuld}. The PQCs operate on qubits, which can exist in a continuous superposition of their basis states, enabling compact representations of complex feature spaces. The model state is encoded in the quantum state of the system, which is progressively transformed through sequences of parameterised quantum gates. Gate parameters are optimised analogously to the trainable weights of a classical neural network using gradient based methods, with the objective of minimising a task specific loss function defined over the circuit outputs. The dimensionality of the state space of quantum systems scales exponentially with the number of qubits, allowing high dimensional functions to be expressed using relatively few qubits \cite{67Hubregtsen}. In combination with entanglement between qubits, this enables suitably designed quantum circuits to achieve highly efficient representations of complex distributions using several orders of magnitude fewer parameters than equivalent classical models \cite{55Abbas}. Moreover, these quantum features can enhance the expressive capacity of the model \cite{68Schuld,55Abbas}, yielding more accurate predictive outputs. QDiffusion-TS directly leverages these advantages through the integration of QNNs into the diffusion architecture.

\subsection{QDiffusion-TS: Quantum Generative Diffusion Model for Time Series}

\begin{figure}[h!]
    \centering
    \includegraphics[width=1\textwidth]{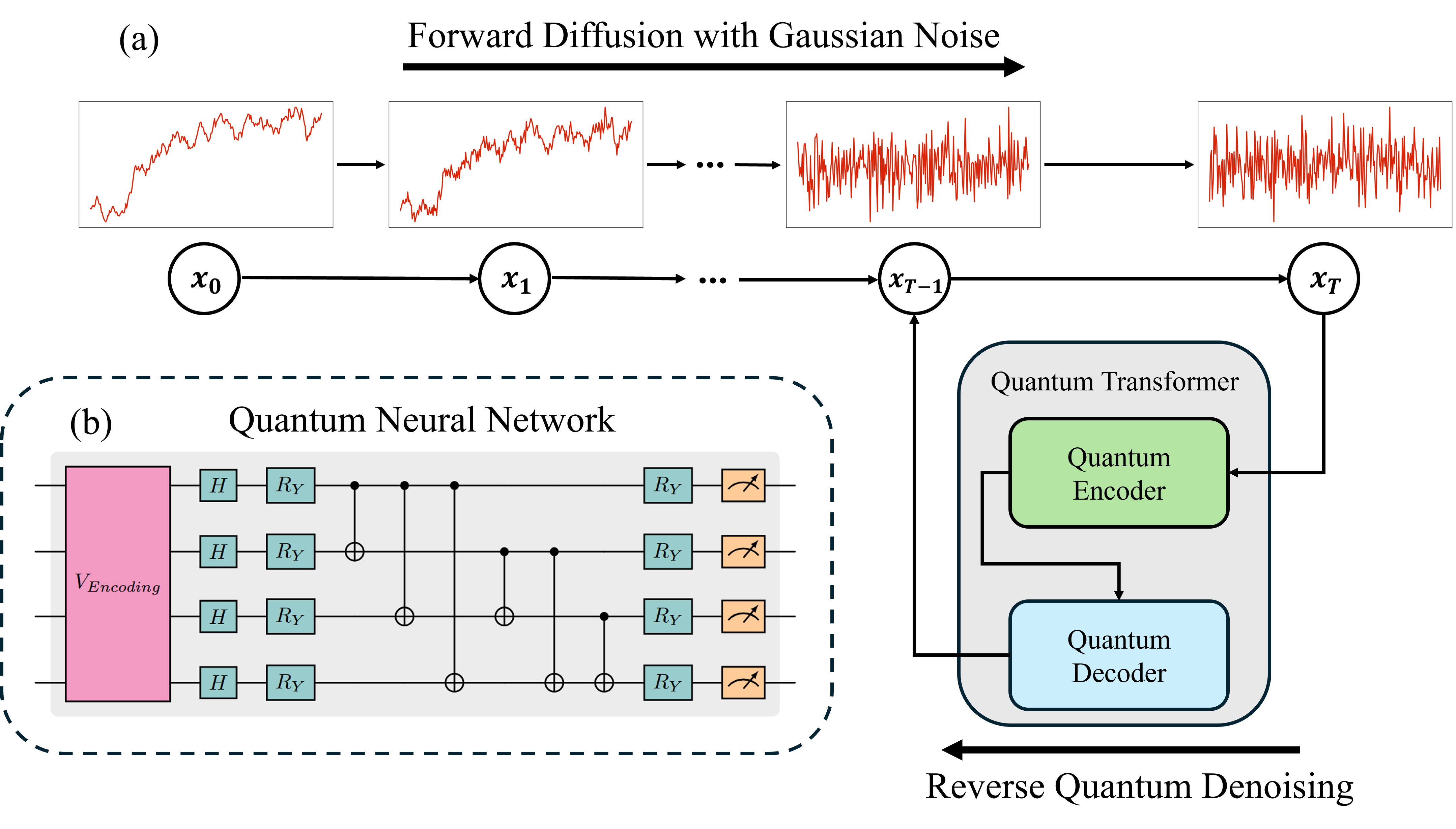}
    \caption{\textbf{a} Schematic of QDiffusion-TS. Noise is progressively added to the original time series over $T$ steps in the forward process, followed by iterative denoising via the quantum enhanced architecture. \textbf{b} Quantum neural network (QNN) architecture utilised within the quantum encoder and decoder blocks. Input data are encoded via the encoding circuit $V_{Encoding}$, processed through a parameterised circuit with controlled rotation gates, and subsequently measured to produce classical outputs.}
    \label{fig:qdiff}
\end{figure}

Diffusion models consist of a forward noising process and a reverse denoising process, in which structured data is progressively transformed into noise and subsequently reconstructed (Fig. \ref{fig:qdiff}a). In the forward process, noise is gradually introduced until the original structure of the data becomes indistinguishable from a simple reference distribution, typically a standard Gaussian. This process is formulated as a fixed analytical Markov chain, in which each noisy state depends only on the preceding state. As a result, a trainable model can be employed to iteratively reverse this process by learning the conditional noise transitions between successive states, progressively removing noise at each step to recover samples from the original data distribution. To generate new samples during inference, a state is initialised from pure noise and refined iteratively through the learned denoising process, producing a unique sample statistically consistent with the underlying data.

To tailor the diffusion model to time series data, the Diffusion-TS framework \cite{19Yuan} employs an encoder-decoder transformer architecture to perform the iterative denoising process, enabling the model to capture both short term dynamics and long range dependencies. This capability arises from the self-attention mechanism \cite{69Vaswani}, which allows transformers to dynamically weight interactions between various time steps, facilitating the learning of complex temporal dependencies across the sequence. Additional modifications include a trend synthetic layer, designed to capture global temporal patterns, and a Fourier synthetic layer, which models both seasonal components and irregular components. Further details of the model architecture are provided in the Methods and Supplementary Information.

To construct QDiffusion-TS, QNNs are systematically incorporated into the reverse denoising process, entirely replacing the feed-forward neural networks (FFNNs) within the transformer encoder and decoder blocks (Fig. \ref{fig:qdiff}b), effectively yielding a quantum transformer architecture for denoising. This substitution reduces the number of parameters per component from 33,088 to just 36, resulting in a reduction approaching three orders of magnitude. When propagated across the full architecture, the replacement of classical feed-forward components throughout the transformer results in a substantial reduction of total model parameters, from 589,030 to 461,254, corresponding  to an overall reduction of approximately 20\%. This section demonstrates that this reduction in model complexity not only preserves the statistical fidelity of the generated time series but in most cases enhances it, as shown through comparison of the quantum enhanced model with the original classical Diffusion-TS framework \cite{19Yuan} (detailed architectural descriptions of both models are provided in Supplementary Materials). For clarity, in the remainder of the study we refer to the QNN enhanced variant as QDiffusion-TS, and compare it against the original classical Diffusion-TS framework.

\subsection{Statistical Properties of Financial Data}
The model is tasked with generating synthetic time series that reproduce the statistical characteristics of real financial data. Financial time series exhibit a set of statistical regularities, commonly referred to as stylised facts, which persist across markets and time periods. Among the most prominent of these are non-Gaussian log-return distributions with heavy tails, and volatility clustering whereby periods of high volatility tend to cluster in time. To assess the statistical fidelity of the generated time series, synthetic samples are compared directly with real data to evaluate the reproduction of these known stylised facts, across both distributional characteristics and temporal dependencies. Real and synthetic log-return distributions are compared using their central moments, in conjunction with quantitative similarity measures, including the Wasserstein distance and Kolmogorov-Smirnov (KS) statistic. Temporal correlations, including volatility clustering, are evaluated via autocorrelation functions, with full details and results provided in the Supplementary Information. These evaluations provide a comprehensive assessment of the model’s capacity to replicate the distributional and dynamical structures of real financial time series.

\subsection{Reproducing Financial Time Series Distributions}
\begin{figure}[h!]
    \centering
    \includegraphics[width=1\textwidth]{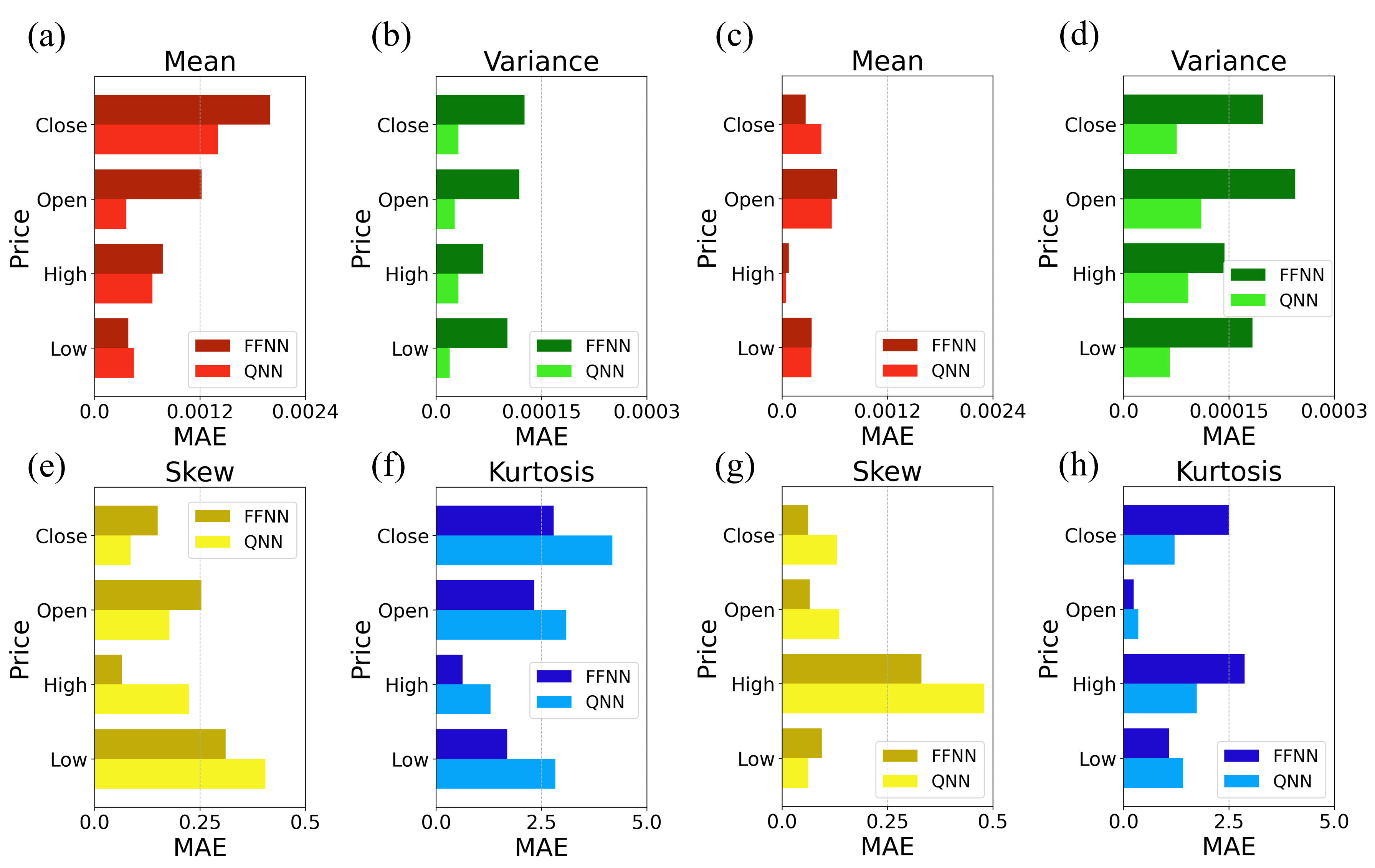}
    \caption{Mean absolute error (MAE) between the synthetic and real central moments of the aggregated log-return distributions for QDiffusion-TS (QNN) and the classical diffusion model (FFNN), evaluated across price features. \textbf{a},\textbf{b} Mean and variance for the Apple dataset. \textbf{c},\textbf{d} Mean and variance for the Amazon dataset. \textbf{e},\textbf{f} Skewness and kurtosis for the Apple dataset. \textbf{g},\textbf{h} Skewness and kurtosis for the Amazon dataset.}
    \label{fig:cent-mom}
\end{figure}

QDiffusion-TS and its classical counterpart were used to generate synthetic financial time series. To evaluate the generative quality of the respective models, the log-return distributions of the synthetic financial time series were compared with those of the real data. Aggregate distributions from the selected datasets were constructed from 2001 generated sequences and 1000 sampled sequences from the real datasets, each comprising of 256 timesteps. The central moments of the generated log-return distributions were evaluated for the Apple and Amazon datasets (Fig. \ref{fig:cent-mom}).
Deviations from the real data were quantified using the mean absolute error (MAE) for both the classical and quantum models. Overall, QDiffusion-TS generally yielded lower MAE than the classical model across the majority of price variables, with the variance reproduced 63\% more accurately on average, while also demonstrating closer agreement in the mean. On the other side, the classical model generally provides a closer approximation of the kurtosis, however the quantum model still captures this feature effectively, indicating that both models are able to reproduce the heavy tailed characteristics of financial returns. The classical model also showed marginally closer agreement in skewness across the two datasets.

\begin{figure}[h!]
    \centering
    \includegraphics[width=\textwidth]{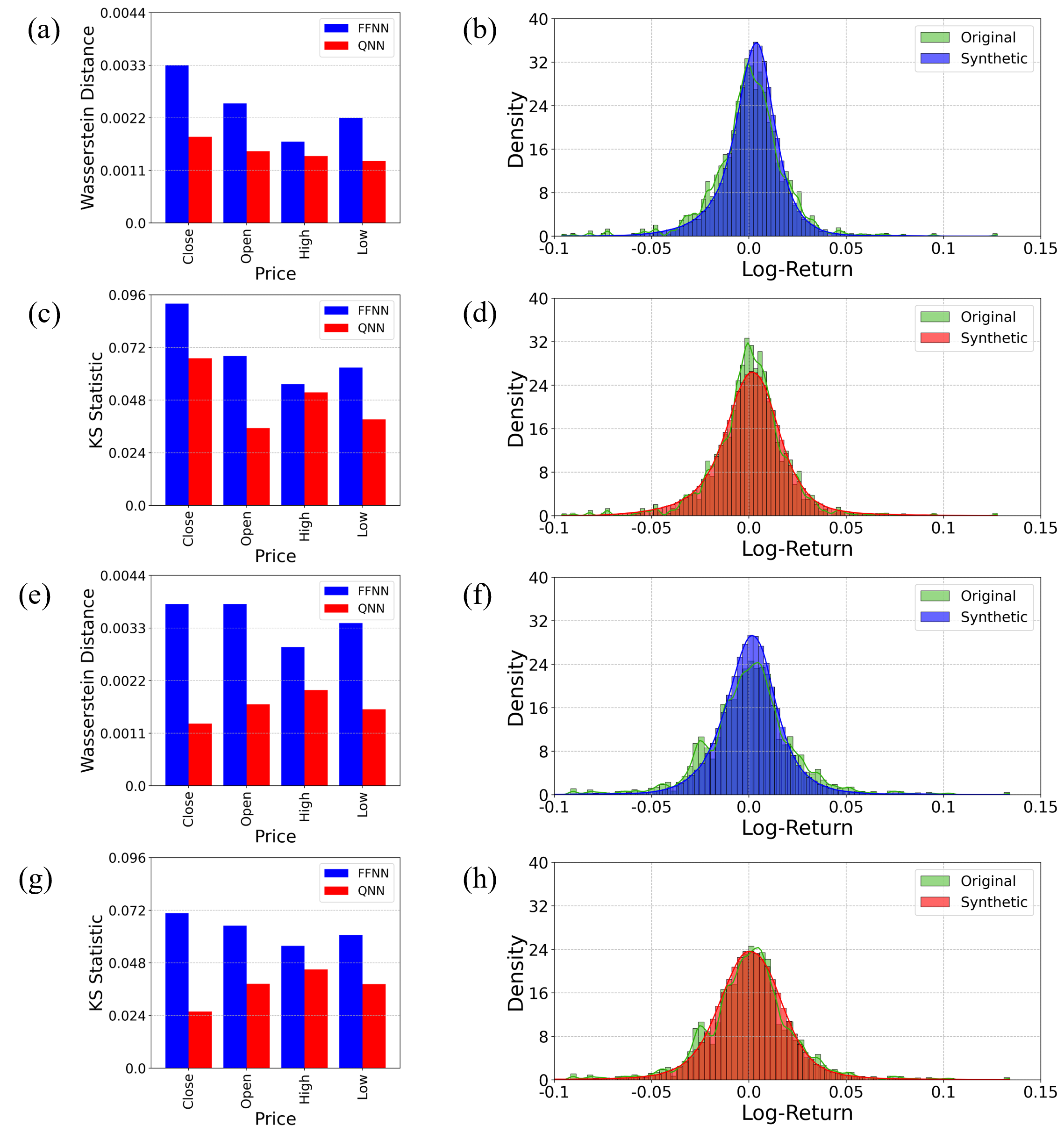}
    \caption{Comparison of aggregated log-return distributions between synthetic and real data for QDiffusion-TS (QNN, red) and the classical diffusion model (FFNN, blue). \textbf{a},\textbf{c} Wasserstein distance and KS statistic between synthetic and real distributions for the Apple dataset. \textbf{b} Estimated probability density function of the Apple open price for real and classical synthetic data. \textbf{d} As in \textbf{b} for quantum synthetic data. \textbf{e}–\textbf{h} Corresponding results for the Amazon dataset.}
    \label{fig:metrics}
\end{figure}

To further quantify the similarity between generated and real log-return distributions, statistical distance metrics were evaluated. Both classical and quantum models reproduce the overall structure of the empirical distributions (Fig. \ref{fig:metrics}b,d,f,h). However, the quantitative comparison indicates that QDiffusion-TS consistently achieves lower statistical distances across both metrics (Fig. \ref{fig:metrics}a,c,e,g). This improvement is more pronounced for the Amazon dataset, which exhibits a 51\% average reduction in Wasserstein distance across features for the quantum model, compared to approximately 36\% for the Apple dataset.

\begin{figure}[h!]
    \centering
    \includegraphics[width=1\textwidth]{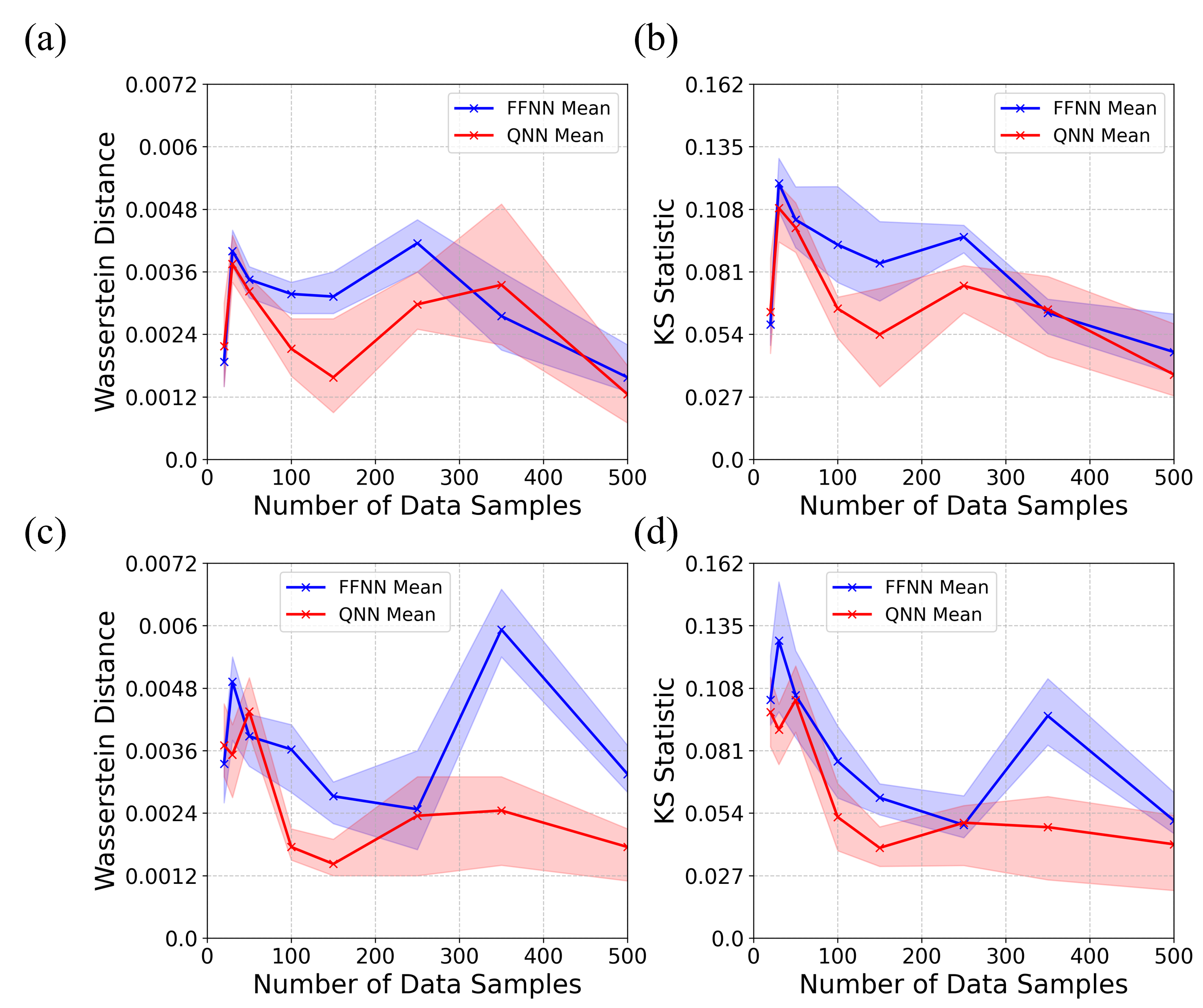}
    \caption{Quantitative comparison of aggregated log-return distributions between synthetic and real data for QDiffusion-TS (QNN) and the classical diffusion model (FFNN) across varying training sample counts. \textbf{a}, \textbf{b} Wasserstein distance and KS statistic between synthetic and real distributions for the Apple dataset. \textbf{c}, \textbf{d} Corresponding results for the Amazon dataset. Values represent mean statistical distances, with error bounds indicating the minimum and maximum across price features.}
    \label{fig:small-data}
\end{figure}

To investigate the effect of training data size on model performance, this analysis was repeated using varying counts of training sequences. The resulting log-return distributions were compared with the real data using the same statistical distance metrics for each training configuration (Fig. \ref{fig:small-data}). At small numbers of training sequences, both models exhibit comparable performance. As the number of training samples increases, the quantum model achieves lower statistical distances in most cases across the range of dataset sizes considered.

\subsection{Downstream Forecasting Task}

\begin{figure}[h!]
    \centering
    \includegraphics[width=\textwidth]{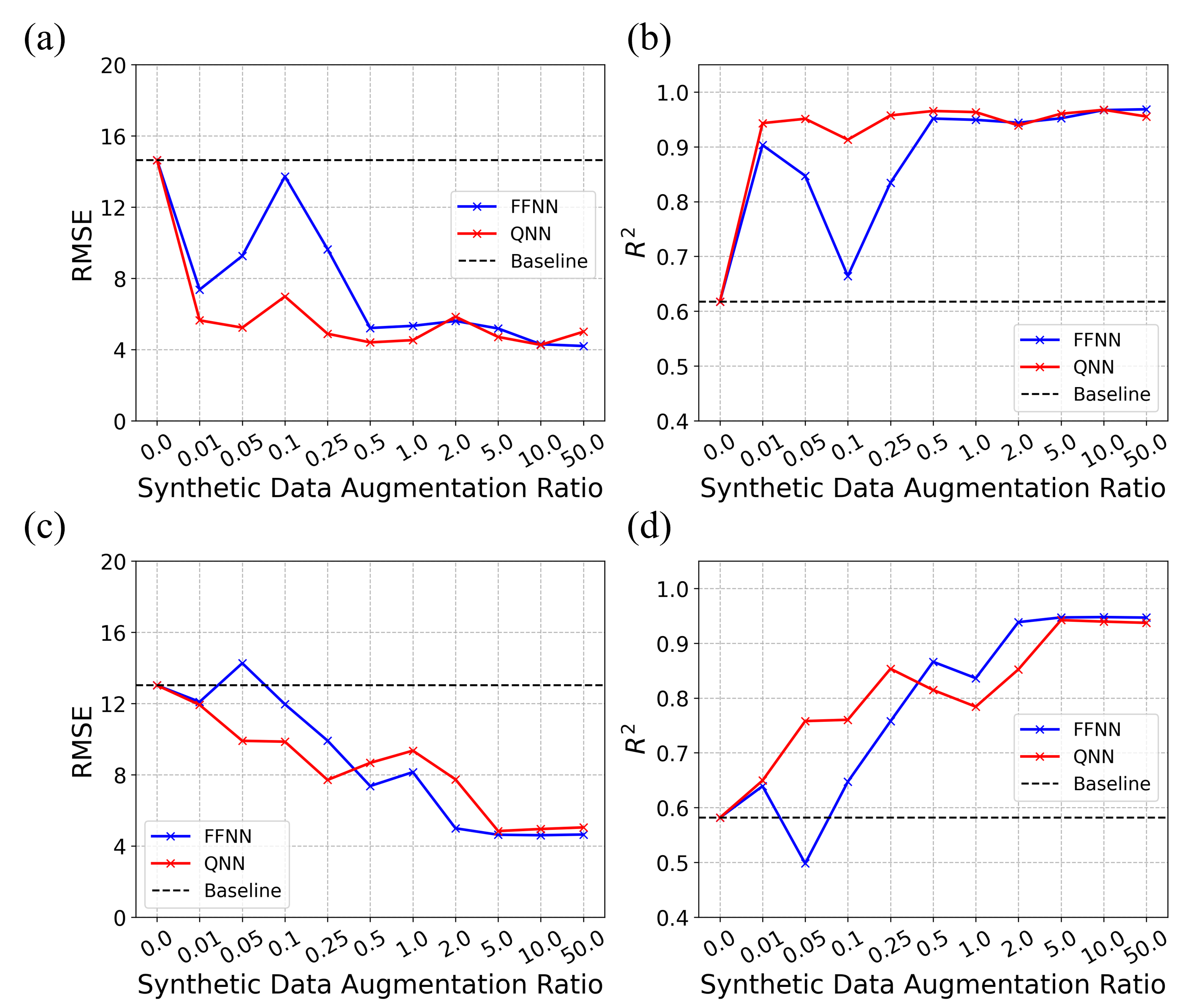}
    \caption{Regression performance metrics for the downstream forecasting task predicting the closing price using training datasets augmented with synthetic data generated by QDiffusion-TS (QNN) and the classical diffusion model (FFNN) at ratios of $1:x$, where $x$ denotes the synthetic data augmentation ratio. \textbf{a}, \textbf{b} RMSE and $R^2$ score between predicted close price and real close price for the Apple dataset. \textbf{c}, \textbf{d} Corresponding results for the Amazon dataset. The baseline corresponds to training without data augmentation. The x-axis is treated as a discrete set of augmentation levels rather than a continuous variable, with each ratio representing an independent experimental configuration.}
    \label{fig:forecast}
\end{figure}

A bidirectional long short-term memory (BiLSTM) model was employed to evaluate the quality and practical utility of synthetic data generated by QDiffusion-TS and its classical counterpart in a forecasting setting, specifically to predict the closing price using a lookback window of 30 trading days (Fig. \ref{fig:forecast}). The training dataset is augmented with synthetic sequences generated using the real financial data at varying ratios of $1:x$, where $x$ denotes the number of synthetic sequences per real sequence. Forecasting performance is assessed using the root mean squared error (RMSE) and coefficient of determination ($R^2$) across these augmentation levels, alongside a baseline model trained solely on real data, to quantify the impact of synthetic data on predictive accuracy.

Augmenting the training data with synthetic sequences generated by both QDiffusion-TS and the classical model leads to a substantial improvement in forecasting performance relative to the baseline across both datasets by up to approximately 71\% in RMSE, demonstrating the practical utility of the generated data for downstream predictive tasks. For the Apple dataset the majority of this improvement is achieved as soon as synthetic data is introduced, with only marginal gains observed as the proportion of synthetic data increases further. In contrast, the Amazon dataset exhibits a more gradual improvement with increasing synthetic data proportion. For both datasets, performances saturates at a ratio 1:5, beyond which further addition of synthetic data yields no improvement. The classical and quantum models achieve highly comparable predictive performance at peak performance, with the quantum model generally outperforming at smaller ratios.

To further contextualise the benefits of synthetic data augmentation, the training set was also expanded using additional real data. Rather than improving performance, this generally degraded it relative to the baseline, with only occasional ratios producing modest gains and none approaching the peak performance achieved through synthetic augmentation. This behaviour is consistent with the non-stationarity of financial time series, where incorporating temporally distant samples introduces distributional shift rather than informative signal (Supplementary Table 3).

\subsection{Quantum Hardware Validation}
\begin{figure}[h!]
    \centering
    \includegraphics[width=\textwidth]{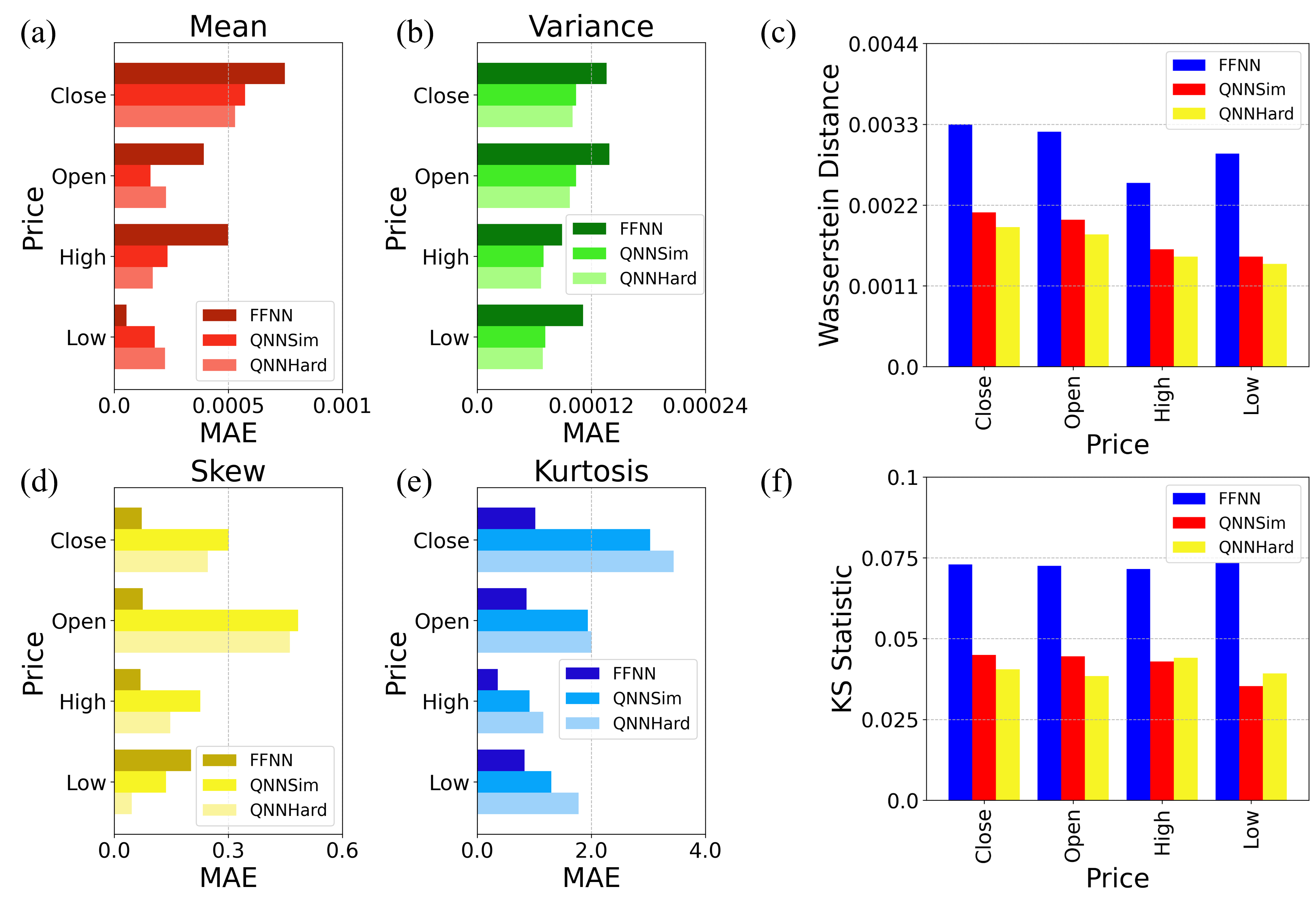}
    \caption{Comparison of aggregated log-return distributions between the classical diffusion model (FFNN), simulated QDiffusion-TS (QNNSim), and hardware executed QDiffusion-TS (QNNHard) for the Apple dataset. \textbf{a},\textbf{b},\textbf{d},\textbf{e} Mean absolute error (MAE) between the synthetic and real central moments evaluated across price features for mean (\textbf{a}), variance (\textbf{b}), skewness (\textbf{d}), and kurtosis (\textbf{e}). \textbf{c},\textbf{f} Wasserstein distance and KS statistic between synthetic and real distributions.}
    \label{fig:qpu}
\end{figure}

Results are further validated on quantum hardware using the IQM Emerald quantum device \cite{79IQM}, where the inference stage of the model is executed to assess the practical feasibility of the proposed approach on current quantum systems. To accommodate hardware constraints, QDiffusion-TS is reduced in capacity while retaining the essential architectural structure, with the classical model adjusted accordingly to ensure a consistent basis for comparison. The simulated quantum model also uses the same reduced architecture as the hardware model, enabling a direct comparison. Distributional comparisons are conducted following the same evaluation procedure as above. Aggregate log-return distributions are constructed using 1224 sampled sequences from the real dataset, 136 sequences generated on quantum hardware, and 2001 sequences from both the classical and simulated quantum models. To ensure compatibility with hardware constraints, the sequence length is reduced to 32 timesteps for all models. The resulting distributions are assessed via their central moments and statistical distance metrics for the Apple dataset (Fig. \ref{fig:qpu}). The results demonstrate that execution on quantum hardware closely matches the performance of the simulated quantum model, with both quantum approaches consistently outperforming the classical model across the statistical distance metrics. In particular, the hardware model produces distributions that more closely align with the real data, reducing Wasserstein distance by approximately 89\% on average across features relative to the classical model. Notably, the hardware implementation also exhibits a marginal advantage over the simulated quantum model, achieving  approximately 8\% lower Wasserstein distance on average across features.

\section{Discussion}
QDiffusion-TS generates synthetic financial time series that more closely match the real data distributions than the equivalent classical diffusion model. Improvements are observed in the central moments of the generated log-return distributions, most notably in variance and, to a lesser extent, mean, indicating a better representation of distributional spread and overall shape. Statistical distances between the generated and real data distributions are lower for the quantum model across both metrics and datasets, with quantum hardware results closely matching simulated performance and, in some cases, marginally surpassing it. When applied to downstream forecasting tasks, the inclusion of synthetic data from both approaches substantially improves predictive performance relative to a baseline trained solely on real data. Notably, the quantum generated sequences achieve equivalent improvements in peak predictive accuracy to those obtained with classical sequences, while requiring significantly fewer trainable parameters. This reduction arises from the direct replacement of the FFNNs within the transformer encoder and decoder blocks with QNNs, producing a hybrid quantum transformer that uses nearly three orders of magnitude fewer trainable parameters in each replaced component, underscoring the parameter efficiency of QDiffusion-TS. This behaviour may be attributed to the high dimensional Hilbert space of the quantum system, in which $n$ qubits encode a state of dimension $2^n$, allowing complex features correlations to be represented with comparatively few parameters. Entangling operations between qubits further enable the efficient representation of strongly correlated features, enhancing the expressive capacity of the model and allowing the quantum neural network to capture distributional structure more effectively than its classical counterpart.

The improved statistical distances observed for the quantum model are primarily driven by its more accurate estimation of variance, which exerts the greatest influence on the overall shape of the log-return distribution. Secondary contributions may arise from improved estimation of the mean, although these effects are comparatively smaller.

Quantum models have been shown to exhibit effective generalisation in low data regimes in certain settings \cite{56Caro}, highlighting potential sample efficiency advantages relative to classical models. However, when evaluated across varying counts of training sequences, both models demonstrate comparable performance at the smallest dataset sizes. Although the quantum model typically achieves lower statistical distances overall, the results do not indicate a distinct advantage specific to the low data regime. These results suggest that the improved performance of the quantum model in this setting is not primarily driven by small data effects, but instead by its ability to more accurately capture the statistical structure of the data distribution. The lack of a pronounced advantage in the smallest data regimes may be attributed to the hybrid nature of the model, in which classical components remain significant, potentially limiting the extent to which quantum advantages in low data regimes can be realised.

Although the quantum model achieves lower statistical distances to the original datasets, predictive performance in the downstream forecasting task is largely similar between the two approaches. This may reflect a complementary balance between fidelity to the source data and variability among generated samples; the quantum model produces sequences that are highly accurate relative to the original data, while the classical model may introduce slightly greater diversity in the synthetic sequences. Both of these characteristics are valuable for the forecasting task, leading to comparable overall predictive performance. A further practical advantage of synthetic augmentation over simply expanding the real training set is that generated sequences are drawn from the learned distribution of the training data, rather than from increasingly distant historical windows. Expanding the real training set was found to degrade performance in most cases, with results fluctuating across expansion ratios and never matching those obtained through synthetic augmentation. This is consistent with the non-stationarity of financial time series where real samples drawn from earlier regimes carry distributional shift relative to the target window, whereas synthetic samples remain anchored to the learned distribution and therefore provide a more stable source of additional training signal (Supplementary Table 3).

The quantum hardware implementation exhibits a slight improvement in statistical accuracy compared to its simulated counterpart. As both models are trained identically using a classical simulator, the only distinction arises during inference, where the hardware executed model is subject to noise inherent to the quantum device. The observed improvement therefore suggests that hardware noise does not degrade performance in this setting and may, in fact, have a beneficial effect. The improved alignment with the target distribution in this setting may be attributed to the additional stochasticity introduced by the quantum hardware increasing variability in the generated samples. It is also notable that the model is not trained under hardware noise conditions, and therefore does not explicitly account for these effects during optimisation. These findings indicate that QDiffusion-TS is robust to noise in near-term quantum hardware, and further suggest that such noise may not solely represent a limitation but could also play a constructive role in certain generative modelling settings, laying the groundwork for a new generation of hybrid classical-quantum generative models.

\section{Methods}

\subsection{Data}
This study uses major publicly traded equities to evaluate the performance of QDiffusion-TS and its classical counterpart. Historical price data for Apple and Amazon (with Microsoft and Google included in the Supplementary Information) were obtained from Yahoo Finance \cite{71Yahoo}. Each dataset comprises open, close, high, and low price features, along with trading volume, over the period 12/02/2021-12/02/2026 at daily temporal resolution.  Models are trained on the log-returns of these time series: 

\begin{equation} \label{eq: log-ret}
    r_i = \ln\left(\frac{y_i}{y_{i-1}}\right)
\end{equation}

\noindent where $y_i$ is the $i$th original data point, and $r_i$ is the respective log-return. All features are standardised prior to training.

For training the diffusion model, each time series is further partitioned into overlapping sequences of fixed length 256. This sampling is taken across the full dataset with a stride of 1, enabling the model to learn local temporal structure. For experiments involving quantum hardware the sequence length is reduced to 32, to accommodate limitations in data encoding capacity. During inference, sequences of the same length are generated. To mitigate sampling bias arising from overlapping windows, a weighted sampling strategy is employed. Each sequence is assigned a sampling probability inversely proportional to the number of times its constituent timesteps appear across all sampled sequences. This reduces the over representation of central regions of the time series and promotes a more uniform contribution of all data points during training.

As the objective of the diffusion model is to reproduce the statistical properties of the data distribution rather than to perform out of sample prediction, it is trained on the full dataset. Evaluation is therefore conducted by comparing the statistical characteristics of generated samples with those of the original data. For the predictive task, strict chronological splitting is applied, and synthetic data are generated using only the training portion of the dataset to prevent look-ahead bias. The data are split in a 59.5:10.5:30 ratio for training, validation, and testing, respectively. 

\subsection{Quantum Neural Network}
To integrate the QNN within QDiffusion-TS for execution on a classical simulator, feature representations must be transferred between the classical transformer blocks and the quantum circuit. To accommodate the limited input size constraints of the QNN, a learnable dimensionality reduction stage is introduced to project the high dimensional feature representations processed by the transformer into a lower dimensional latent space compatible with the quantum circuit. Principal component analysis (PCA) is additionally applied to further compress the feature space whilst preserving the structure of the data. The compressed features are encoded into a quantum state using amplitude encoding, whereby a normalised classical vector is embedded into the amplitudes of the quantum state, permitting the representation of $2^n$ features using $n$ qubits. 
Accordingly, a 9-qubit system is employed, enabling the encoding of 512 inputs. This corresponds to a sequence length of 256 with two features, obtained by applying PCA to reduce the original five features of the dataset. The ansatz employed is the 'real amplitudes' circuit for remote sensing imagery classification \cite{72Sebastianelli}, and consists of an initial layer of Hadamard gates, followed by parameterised $R_y$ gates with independent trainable parameters, and CNOT entangling operations (Fig. \ref{fig:qdiff}b). The variational block is repeated twice in order to enhance the expressive capacity of the model. Leveraging the direct access to the simulated quantum state provided by the classical simulator, the full statevector amplitudes are extracted as the QNN output. The resulting vector is subsequently projected back to the original transformer dimensions before being passed to the remaining layers of the model. All classical simulations are performed using the Pennylane Python library.

For execution on quantum hardware, this model is adapted to accommodate the constraints of near-term quantum hardware with modified encoding and measurement methods, alongside a hardware efficient ansatz (Supplementary Information). In contrast to classical simulation, direct amplitude encoding and full statevector readout are not accessible on hardware platforms. To address this, classical data are encoded using angle encoding \cite{73Schuld}, where each qubit encodes a feature via rotations about the y-axis. The resulting quantum state is given by

\begin{equation}
    \ket{\psi(\textbf{x})} = \bigotimes^{N}_{i=1}R_y(x_{i})\ket{0}
\end{equation}

\noindent for $N$ qubits and $N$ data points $\textbf{x} = (x_1, \dots, x_N)$. Similarly, measurement is performed by evaluating the expectation values of the Pauli-Z operators on each qubit using 128 measurement shots. For a given qubit $i$, this is given by

\begin{equation}
\left <\sigma_Z^{(i)} \right > = \bra{\psi(\mathbf{x})} \sigma_Z^{(i)} \ket{\psi(\mathbf{x})}
\end{equation}

\noindent and the expectation values are concatenated to form the classical output vector. To ensure a direct comparison, the simulated quantum model used in the hardware experiments adopts the same hardware compatible architecture. Quantum computations are executed on the IQM Emerald device via Amazon Braket. Due to the computational cost of iterative training on current quantum hardware, only the inference stage is performed on the real quantum hardware, while model training is conducted using the simulator of the quantum device.

\subsection{Quantum Generative Diffusion for Time Series}
The forward diffusion process, following the formulation of Diffusion-TS \cite{19Yuan}, progressively corrupts a sample from the data distribution $x_0 \sim q(x_0)$ through the iterative addition of noise until the state approaches an isotropic Gaussian distribution $x_T \sim \mathcal{N}(0, \textbf{I})$. This transition is defined by the conditional distribution 

\begin{equation}
    q(x_t|x_{t-1})=\mathcal{N}(x_t;\sqrt{1-\beta_t}x_{t-1}, \beta_t\textbf{I})
\end{equation} 

\noindent which forms a Markov chain over T timesteps;

\begin{equation}
    q(x_{1:T}|x_0) = \prod_{t=1}^Tq(x_t|x_{t-1})
\end{equation}
\noindent where $\beta_t \in (0,1)$ defines the variance schedule controlling the magnitude of noise added at each timestep, chosen to increase gradually with $t$ to ensure a smooth transition. Conventionally, diffusion models parameterise the reverse process $p_\theta(x_{t-1}|x_t)$ by predicting the noise component added at each timestep, thereby enabling progressive denoising. In contrast, here the architecture directly predicts an estimate of the original data sample $\hat{x}_0(x_t, t, \theta)$, using the quantum transformer. Thus, the standard noise prediction objective is reparameterised as, 

\begin{equation}
    \mathcal{L}=\mathbb{E}_{t, x_0}\left[\omega_t\|x_0-\hat{x}_0(x_t, t, \theta)\|^2\right]
\end{equation}

\noindent where the weighting term $\omega_t$ accounts for the noise at timestep t, defined by

\begin{equation}
    \omega_t=\frac{\lambda\sqrt{\alpha_t}\sqrt{(1-\bar{\alpha}_t)}}{\beta_t}
\end{equation}

\noindent with $\alpha_t=1-\beta_t$, $\bar{\alpha}_t=\displaystyle\prod_{s=1}^t\alpha_s$ and $\lambda$ is a constant scaling factor. This weighting emphasises reconstruction at higher noise levels, where the denoising task is more challenging. In addition, a Fourier-based loss term is introduced \cite{19Yuan} using the Fourier transform, to encourage accurate reconstruction of the frequency characteristics of the time series and improve recovery of temporal structure. The resulting loss function can be written as,

\begin{equation}
    \mathcal{L}_{Fourier}=\mathbb{E}_{t,x_0}\left[\omega_t\left[\lambda_1\|x_0-\hat{x}_0(x_t, t, \theta)\|^2+\lambda_2\|\mathcal{FFT}(x_0)-\mathcal{FFT}(\hat{x}_0(x_t, t, \theta))\|^2\right]\right]
\end{equation}

\noindent where $\mathcal{FFT}$ denotes the Fast Fourier Transform and $\lambda_1, \lambda_2$ are weighting coefficients to balance the contributions of the loss terms.

\subsection{Evaluation Metrics}
The log-returns of both real and synthetic data are used when comparing distributions, consistent with standard practice in financial modelling. To assess the accuracy of the central moments of the generated distributions, the MAE is reported. This measures the average absolute deviation between the estimated moment $\hat{x_i}$ and the corresponding value $x_i$ across $N$ samples:

\begin{equation}
    MAE = \frac{1}{N}\sum_{i=1}^N \left| \hat{x_i} - x_i \right|.
\end{equation}

To evaluate similarity between the synthetic and real distributions, the Wasserstein distance and KS statistic are used as quantitative measures. The Wasserstein distance \cite{75Kantorovich} quantifies the minimal cost to transform one distribution into another through the transportation of probability mass, accounting for both the magnitude of the mass and the distance which it is moved. The Wasserstein distance between two distributions $u$ and $v$ is defined as:

\begin{equation}
    l_1(u, v) = \inf_{\pi \in \Gamma(u, v)}\int_{\mathbb{R}\times\mathbb{R}}|x-y|\, d\pi(x,y),
\end{equation}

\noindent where $\Gamma(u,v)$ is the set of distributions on $\mathbb{R} \times \mathbb{R}$ whose marginals are $u$ and $v$, respectively. The KS statistic \cite{76Massey} provides a measure of distributional similarity by evaluating the maximum absolute difference between the cumulative distributions functions of the two samples.

In order to evaluate the predictive performance of the downstream forecasting task, standard regression metrics are employed. RMSE measures the average magnitude of the prediction errors, giving greater weight to larger deviations, defined as:

\begin{equation}
RMSE = \sqrt{\frac{1}{N} \sum_{i=1}^N \left( \hat{x_i} - x_i \right)^2 }.
\end{equation}

The $R^2$ score measures the fraction of the total variance in the observed data that is accounted for by the predictions, defined as:

\begin{equation}
R^2 = 1 - \frac{\sum_{i=1}^N \left( x_i - \hat{x_i} \right)^2}{\sum_{i=1}^N \left( x_i - \bar{x} \right)^2}
\end{equation}

\noindent where $\hat{x}$ denotes the mean of the true values. An $R^2$ score of 1 indicates perfect predictive accuracy, while a score of 0 indicates that the model performs no better than simply predicting the mean of the observed data.

\backmatter

\bmhead{Supplementary information}
%you'll need to compile once to generate the .bbl file, then paste its contents into your .tex file in place of the \bibliography{sn-bibliography}+line
Supplementary Information is available for this paper.

\bmhead{Acknowledgements}

The authors acknowledge the UK National Quantum Computing Centre (NQCC) for providing access to quantum hardware through the SparQ programme. J.W. acknowledges support from the Kingston University PhD Studentship.

\section*{Declarations}

\subsection*{Funding}
This project was partially funded and supported by the UK National Quantum Computer Centre [NQCC200921], which is a UKRI Centre and part of the UK National Quantum Technologies Programme (NQTP).

\subsection*{Competing interests}
The authors declare no competing interests.

\subsection*{Ethics approval and consent to participate}
Not applicable

\subsection*{Consent for publication}
Not applicable

\subsection*{Data availability }
The datasets used in this study are available at https://finance.yahoo.com.

\subsection*{Materials availability}
Not applicable

\subsection*{Code availability }
Code is available from the corresponding author upon reasonable request and will be made publicly available on GitHub upon publication.

\subsection*{Author contribution}
J.W., X.L., and F.C. conceived and designed the study. J.W. developed the model, implemented the code, conducted the experiments, analysed the results, generated the figures, and wrote the original draft. X.L., D.M., R.N. and F.C. supervised the project. X.L. and D.M. secured funding. X.L., F.C., D.M., and R.N. reviewed and edited the manuscript.


%\bibliography{bibliography}% common bib file

\end{document}